\documentclass{article} 
\usepackage{iclr2026_conference,times}

\usepackage{graphicx}
\usepackage{url}
\usepackage[dvipsnames]{xcolor}
\usepackage{xspace}
\usepackage{amssymb}
\usepackage{amsmath}
\usepackage[colorlinks=true]{hyperref} 
\usepackage[capitalize]{cleveref} 
\usepackage{algorithm}
\usepackage{algorithmicx}
\usepackage{algpseudocode}
\usepackage{wrapfig}
\usepackage{booktabs}
\usepackage{multirow}
\usepackage[table]{xcolor}

\hypersetup{
    citecolor=YellowGreen,     
}

\definecolor{catgray}{gray}{0.92}

\algrenewcommand\algorithmicindent{1.2em}

\crefname{section}{Sec.}{Secs.}
\Crefname{section}{Sec.}{Secs.}
\crefname{algorithm}{Alg.}{Algs.}
\Crefname{algorithm}{Alg.}{Algs.}

\makeatletter
\DeclareRobustCommand\onedot{\futurelet\@let@token\@onedot}
\def\@onedot{\ifx\@let@token.\else.\null\fi\xspace}

\def\ie{\emph{i.e}\onedot}

\def\wrt{w.r.t\onedot}

\makeatother

\usepackage{amsmath,amsfonts,bm}

\def\eqref#1{equation~\ref{#1}}

\def\1{\bm{1}}

\DeclareMathAlphabet{\mathsfit}{\encodingdefault}{\sfdefault}{m}{sl}
\SetMathAlphabet{\mathsfit}{bold}{\encodingdefault}{\sfdefault}{bx}{n}

\title{Rolling Forcing: Autoregressive Long Video Diffusion in Real Time}

\author{
    \textbf{Kunhao Liu}$^1$\thanks{Work done during internship at ARC Lab, Tencent PCG.}~~
    \textbf{Wenbo Hu}$^2$\thanks{Corresponding authors.}~~
    \textbf{Jiale Xu}$^2$~~
    \textbf{Ying Shan}$^2$~~
    \textbf{Shijian Lu}$^{1\dagger}$\\
    $^1$Nanyang Technological University~~
    $^2$ARC Lab, Tencent PCG
}

\iclrfinalcopy 
\begin{document}

\maketitle

\begin{figure}[h]
\centering
\vspace{-1em}
\includegraphics[width=\textwidth]{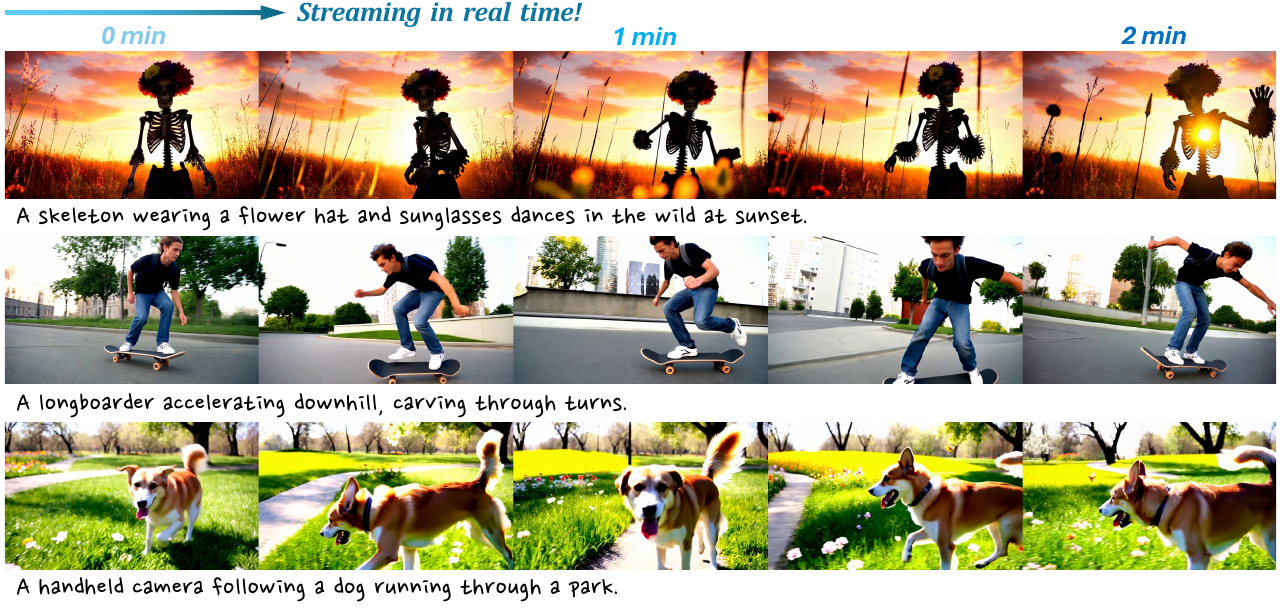}
\vspace{-1em}
\caption{Rolling Forcing performs real-time streaming text-to-video generation at 16 fps on a single GPU and is capable of producing multi-minute-long videos with minimal error accumulation. 
More results, code, and demo can be found at the
\href{https://kunhao-liu.github.io/Rolling_Forcing_Webpage/}{project page}.
}
\label{fig:teaser}
\end{figure}

\begin{abstract}
Streaming video generation, as one fundamental component in interactive world models and neural game engines, aims to generate high-quality, low-latency, and temporally coherent long video streams. However, most existing work suffers from severe error accumulation that often significantly degrades the generated stream videos over long horizons.
We design Rolling Forcing, a novel video generation technique that enables streaming long videos with minimal error accumulation. 
Rolling Forcing comes with three novel designs.
First, instead of iteratively sampling individual frames, which accelerates error propagation, we design a joint denoising scheme that simultaneously denoises multiple frames with progressively increasing noise levels. This design relaxes the strict causality across adjacent frames, effectively suppressing error growth.
Second, we introduce the attention sink mechanism into the long-horizon stream video generation task, which allows the model to keep key–value states of initial frames as a global context anchor and thereby enhances long-term global consistency.
Third, we design an efficient training algorithm that enables few-step distillation over largely extended denoising windows. This algorithm operates on non-overlapping windows and mitigates exposure bias conditioned on self-generated histories.
Extensive experiments show that Rolling Forcing enables real-time streaming generation of multi-minute videos on a single GPU, with substantially reduced error accumulation.

\end{abstract}
\section{Introduction}

Modern video diffusion models~\citep{openai_sora, polyak2024movie} have demonstrated impressive capabilities in generating short video clips with rich detail and coherent motion.
However, interactive applications such as world models~\citep{bruce2024genie}, neural game engines~\citep{valevski2024diffusion}, and immersive XR environments require the ability to \emph{stream} each frame with \emph{minimal latency} while maintaining visual quality and temporal coherence over \emph{long horizons}.
Unlike offline video generation, where the entire sequence is synthesized together at one go, the streaming video generation operates in an online fashion: frames are generated sequentially and immediately consumed by downstream tasks or displayed to users.
Such online nature imposes unique challenges, as the model must maintain long-horizon consistency while accommodating real-time constraints in an autoregressive manner.

\begin{figure}[t]
\centering
\includegraphics[width=\textwidth]{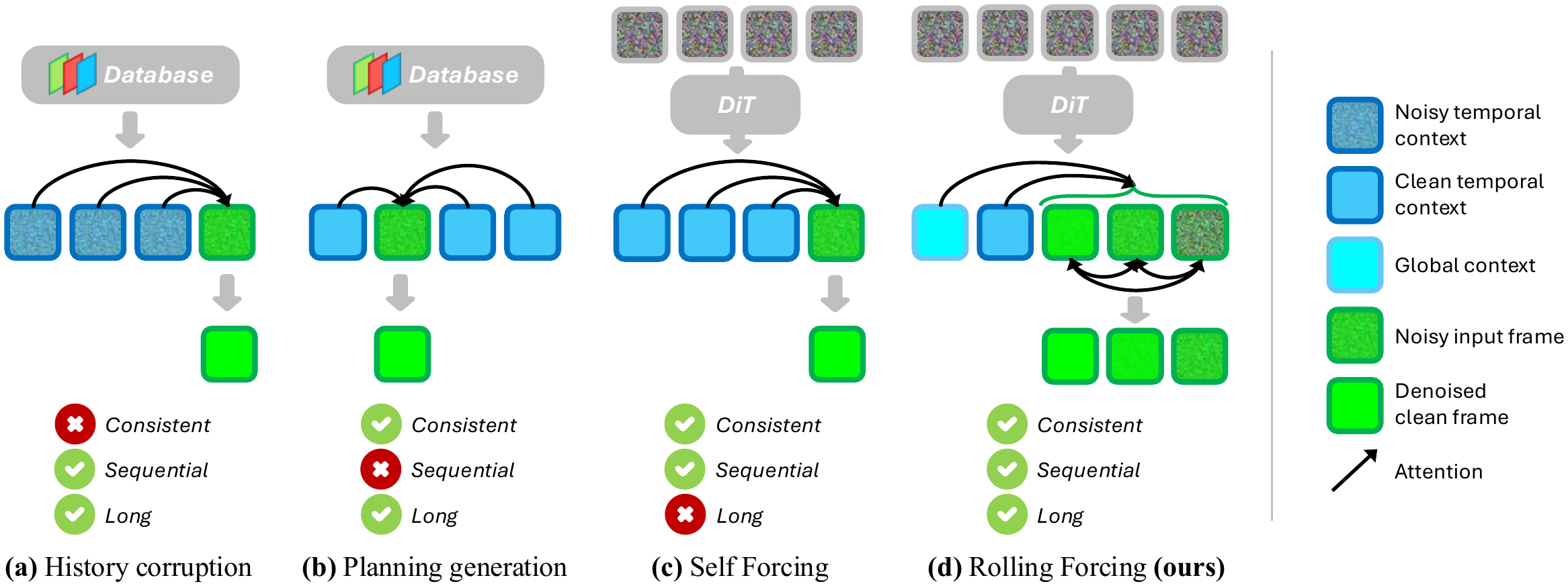}
\caption{Different paradigms in autoregressive video generation. History corruption~\citep{chen2024diffusion, guo2025long} in (a) compromises temporal consistency, while planning generation~\citep{zhang2025packing, xiang2025macro} in (b) is incompatible with sequential streaming video generation. Self Forcing~\citep{huang2025self} in (c) can achieve consistent sequential streaming but suffers from severe error accumulation while generating long videos. The proposed Rolling Forcing in (d) supports streaming long video generation with superior temporal consistency and minimal error accumulation.}
\label{fig:intro}
\end{figure}

Real-time streaming video generation methods, such as CausVid~\citep{yin2025slow} and Self Forcing~\citep{huang2025self} (illustrated in \cref{fig:intro}(c)), distill a pretrained bidirectional video diffusion model into a fast, causal autoregressive generator. While they enable consistent sequential generation, their strictly causal frame prediction causes each frame to inherit errors from its predecessors, allowing small imperfections to compound over long horizons and eventually leading to noticeable drift and quality degradation.
Two representative approaches have been explored for improving video generation over long horizons, as illustrated in \cref{fig:intro}(a,b). The first approach explores \emph{history corruption}, which injects noise into past frames to reduce over-reliance on histories~\citep{chen2024diffusion, guo2025long}. History corruption mitigates drift by narrowing the gap between self-generated and ground-truth context, but it deprives the model of clean references and compromises temporal consistency. The second approach explores \emph{planning generation} by first synthesizing distant key frames and then interpolating intermediates~\citep{zhang2025packing, xiang2025macro}. Anchoring distant frames to the initial context mitigates drift, but the introduced out-of-order schedule violates strict sequential emission, which is unsuitable for real-time streaming.

We design \emph{Rolling Forcing}, an autoregressive long video generation technique that mitigates error accumulation while maintaining real-time performance as illustrated in \cref{fig:intro}. 
Rolling Forcing comes with three new designs. 
First, instead of iteratively denoising a single frame at a time as in most existing work, Rolling Forcing introduces rolling-window denoising to process multiple consecutive frames simultaneously. Within each window, frames are connected by bidirectional attention and assigned progressively increasing noise levels. Such mutual refinement corrects local errors before any frame is finalized, thereby suppressing long-horizon drift. In addition, this design allows us to emit a clean frame after each single forward pass, achieving real-time throughput on a single GPU despite a much larger attention window. 
Second, we adapt the attention sink mechanism~\citep{xiao2023efficient} to the streaming video generation task, thereby strengthening long-term global consistency. Specifically, we persist the key–value states of the initial frames as a global context anchor and dynamically adjust their Rotary Position Embeddings (RoPE)~\citep{su2024roformer}, which freezes the relative positions of initial frames to the current denoising frames and prevents excessive offsets. Note that the KV caching is applied to the recent clean frames as well to reduce latency and maintain temporal consistency. 
Third, we design an efficient training algorithm that enables few-step distillation over the extended denoising windows. This algorithm operates on non-overlapping windows that collectively cover all video frames, mitigating exposure bias by conditioning on self-generated histories during training. Extensive experiments show that Rolling Forcing achieves real-time streaming generation of multi-minute videos on a single GPU, with substantially reduced error accumulation as illustrated in \cref{fig:teaser}.

The contributions of this work can be summarized in three key aspects. \textit{First}, we introduce a rolling-window joint denoising technique that processes multiple frames in a single forward pass, enabling mutual refinement while preserving real-time latency. \textit{Second}, we introduce the attention sink mechanism into the streaming video generation task, a pioneering effort that enables caching the initial frames as consistent global context for long-term coherence in video generation. \textit{Third}, we design an efficient training algorithm that operates on non-overlapping windows and conditions on self-generated histories, enabling few-step distillation over extended denoising windows and concurrently mitigating exposure bias.

\section{Related Work}

\paragraph{Bidirectional Video Generation Models.}
Video generation has advanced rapidly in recent years, with modern approaches mostly adopting the paradigms of denoising diffusion. Video diffusion has been explored in both pixel space~\citep{ho2022imagen, singer2022make} and latent space~\citep{blattmann2023align, blattmann2023stable}, with architectures evolving from early Space–Time U-Nets~\citep{blattmann2023stable, hong2022cogvideo} to more recent DiT-based designs~\citep{peebles2023scalable, gupta2024photorealistic}. Significant industrial investment has driven the development of large video diffusion models, leading to several multi-billion parameter models, including open-source models such as Wan~\citep{wan2025wan} and Hunyuan~\citep{kong2024hunyuanvideo}, and closed-source models such as Sora~\citep{openai_sora}, Movie-Gen~\citep{polyak2024movie}, and Seaweed~\citep{seawead2025seaweed}. Notably, these models operate as bidirectional video diffusion models, as they have access to both past and future frames during denoising. While this bidirectional context enables high-quality synthesis for offline generation, it is incompatible with the causality that is necessitated in real-time streaming video generation.

\paragraph{Autoregressive Video Generation Models.}
To enable long video generation, several studies have extended the generation paradigm from bidirectional to autoregressive, which naturally supports gradual rollout over extended time horizons. Autoregressive models are typically trained with next-token prediction objectives and generate spatiotemporal tokens sequentially at inference time~\citep{bruce2024genie, kondratyuk2023videopoet, wang2024loong, weissenborn2019scaling, yan2021videogpt}. More recently, a separate line of research combines autoregressive modeling with denoising diffusion~\citep{chen2024diffusion, gu2025long, guo2025long, jin2024pyramidal, li2024arlon, liu2024mardini, weng2024art, yin2025slow, zhang2025test, zhang2025packing, huang2025self, henschel2025streamingt2v}, where frames are generated one-by-one in an outer loop and each frame is gradually denoised in an inner loop. Within this family, Rolling Diffusion~\citep{ruhe2024rolling} and its variants~\citep{kim2024fifo, teng2025magi, sun2025ar, xie2025progressive, chen2025skyreels, teng2025magi} merge the outer and inner loops: the diffusion model jointly denoises multiple frames at progressively increasing noise levels. However, these methods mostly suffer from exposure bias and error accumulation when generating long videos. Another line of research addresses error accumulation with planning generation~\citep{long2024videostudio, zhao2024moviedreamer, hu2024storyagent, xie2024dreamfactory, zhang2025packing, bansal2024talc, yang2024synchronized, xiang2025macro}, which predicts distant future frames first and then interpolates the intermediate frames. While effective for reducing drift, it breaks the strict sequential order required for real-time streaming. In contrast, our work enables much longer streaming video generation with minimal error accumulation while addressing exposure bias.

\paragraph{Concurrent and Closed-Source Work.} 
Two concurrent works are also devoted to the streaming generation of long videos.
Specifically, StreamDiT~\citep{kodaira2025streamdit} adopts the FIFO-style denoising~\citep{kim2024fifo} for streaming video generation. It modifies the pretrained model architecture by introducing micro-steps and window attention, necessitating extensive additional pretraining with large-scale data and computation. In contrast, our method keeps the pretrained model architecture unchanged, can be trained efficiently in only 3,000 steps, and does not require any video data.
APT2~\citep{lin2025autoregressive} instead explores adversarial distillation~\citep{lin2025diffusion} for streaming video generation. It denoises videos block-by-block and involves multiple costly post-training stages, including diffusion adaptation, consistency distillation, adversarial training, and long-video training. APT2 is trained on one-minute-long videos, whereas ours is trained only on 5-second clips, yet can extend to multi-minute sequences during inference.
Note that both StreamDiT and APT2 are closed-source and trained based on internal video diffusion models (Movie-Gen~\citep{polyak2024movie} and Seaweed~\citep{seawead2025seaweed}), 
while our model is trained on public datasets and relies on an open-source model (\ie, Wan2.1~\citep{wan2025wan}) as its foundation.

\section{Methods}

\subsection{Preliminaries: Exposure Bias in Autoregressive Video Diffusion Models}
An autoregressive video diffusion model is a hybrid generative framework that integrates autoregressive chain-rule decomposition with denoising diffusion for video generation.
Formally, given a sequence of $N$ video frames $x^{1:N} = (x^1, x^2, \dots, x^N)$, their joint distribution can be factorized using the chain rule: $p(x^{1:N}) = \prod_{i=1}^{N} p(x^i \mid x^{<i}).$ Each conditional distribution $p(x^i \mid x^{<i})$ is modeled through a diffusion process, where each frame is generated by progressively denoising Gaussian noise while conditioning on the previously generated frames. 
In practice, one may also generate a chunk of consecutive frames instead of a single frame at each step~\citep{yin2025slow, teng2025magi, huang2025self}. For clarity, we refer to each chunk simply as a frame in the following text.

Autoregressive video diffusion models are trained either (1) from scratch with frame-wise denoising loss or (2) by distilling a pretrained bidirectional model. The first approach is trained under the paradigm of Teacher Forcing (TF) or Diffusion Forcing (DF)~\citep{chen2024diffusion}. In TF, the conditional distribution for the $i$th frame at noise level $t_j$ is $p(x^i_{t_j} \mid x^{<i}_0)$, where all conditional history frames are the ground-truth clean frames from the training data. While in DF, the conditional distribution is $p(x^i_{t_j} \mid x^{<i}_{t_{\ge0}})$, where the history frames are the ground-truth frames corrupted with independent noise levels. Since training relies on ground-truth histories while inference relies on the model’s own predictions, a train–test gap known as exposure bias arises~\citep{schmidt2019generalization}. Mitigating the exposure bias is difficult because the denoising loss requires pairs of model predictions and the corresponding ground truth conditioned on them, which are unavailable. 

The second approach of distillation, however, provides a way to bypass the denoising loss and mitigate exposure bias. CausVid~\citep{yin2025slow} distills a pretrained bidirectional model into a few-step causal model. It adopts a Distribution Matching Distillation (DMD) loss~\citep{yin2024one} that minimizes the reverse KL divergence across randomly sampled timesteps $t$ between the smoothed data distribution $p_{\text{data}}(x_{t})$ and the student generator's output distribution $p_{\text{gen}}(x_{t})$. 
The gradient of the reverse KL can be approximated as the difference between two score functions:
\begin{multline}
\label{eq:dmd}
\nabla_{\theta} \mathcal{L}_{\text{DMD}} \triangleq \mathbb{E}_{t} \left( \nabla_{\theta} \text{KL} \left( p_{\text{gen}, t} \| p_{\text{data}, t} \right) \right) \\
 \approx - \mathbb{E}_{t} \left( \int \left( s_{\text{data}} \left( \Psi \left( G_{\theta}(\epsilon), t \right), t \right) - s_{\text{gen}} \left( \Psi \left( G_{\theta}(\epsilon), t \right), t \right) \right) \frac{d G_{\theta}(\epsilon)}{d \theta} \, d\epsilon \right), 
\end{multline}
where $\Psi$ represents the forward diffusion process, $\epsilon$ is random Gaussian noise, $G_{\theta}$ is the generator parameterized by $\theta$, and $s_{\text{data}}$ and $s_{\text{gen}}$ represent the score functions trained on the data and generator's output distribution, respectively. Since training with DMD loss does not require ground-truth image or video data~\citep{yin2024improved}, Self Forcing~\citep{huang2025self} mitigates the exposure bias by conditioning each frame on previously self-generated histories during training. However, although exposure bias is alleviated, severe error accumulation still occurs once generation extends beyond the trained temporal window.

\subsection{Autoregressive Video Generation via Rolling Diffusion Window}
\label{sec:inference}

\begin{figure}[t]
\centering
\vspace{-1em}
\includegraphics[width=\textwidth]{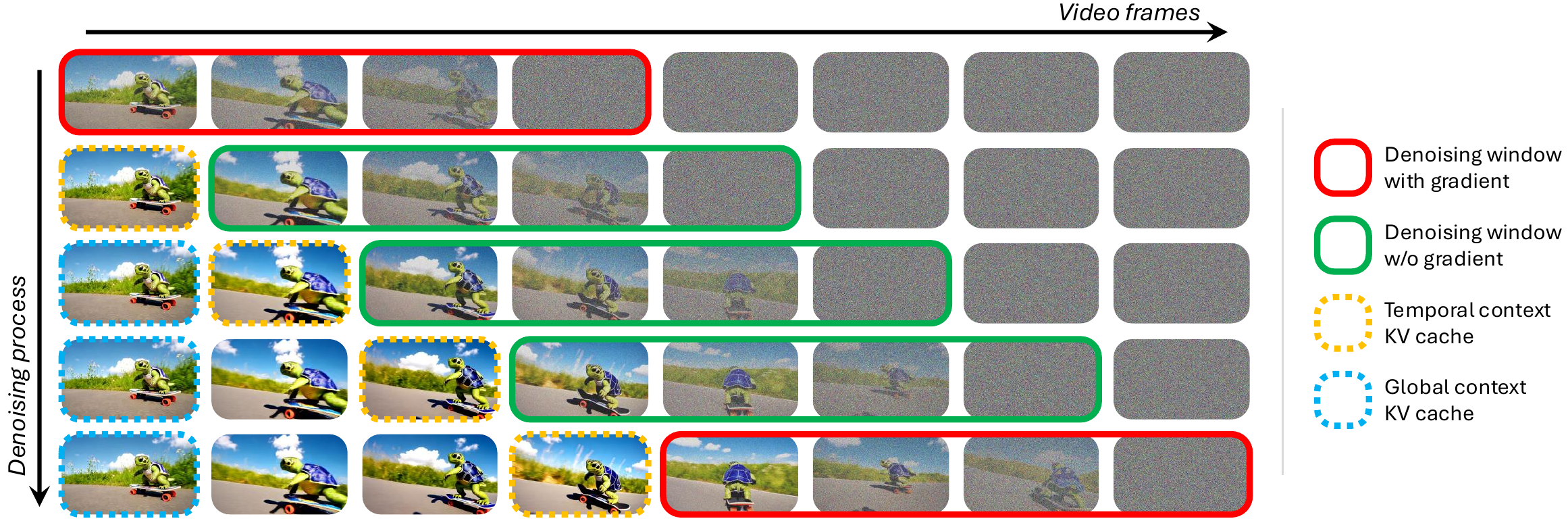}
\caption{Illustration of the Rolling Forcing denoising process with $T=4$. Rolling Forcing jointly denoises a short window of consecutive frames that are assigned progressively higher noise levels and connected by bidirectional attention. The KV cache of recent frames is preserved as temporal context to maintain short-term consistency, while the KV cache of the initial frames is preserved as global context to ensure long-term consistency. During training, only a subset of denoising windows requires gradient computation, as highlighted by the red windows. These windows are mutually exclusive yet collectively cover all video frames.}
\label{fig:method}
\end{figure}

In Self Forcing (SF), videos are generated frame-by-frame in a strict causal manner. Consider a noise schedule $\{t_0=0,t_1, \dots, t_{T}=1000\}$ with total noise levels $T+1$. At each denoising step $t_j$ and frame index $i$, the model denoises an intermediate noisy frame \( x^i_{t_j} \) conditioned on previous clean frames \( x^{<i}_0 \) and then injects Gaussian noise with a lower noise level into the predicted denoised clean frame via the forward diffusion process $\Psi$. This produces a noisy frame \( x^i_{t_{j - 1}} \) which will be used as the input to the next denoising step. Formally, in SF, the denoising process is achieved by: $ x^i_{t_{j - 1}} = \Psi\big(G_\theta(x^i_{t_j}, t_{j},  x^{<i}_0), t_{j-1}\big)$, \mbox{and $ x^i_{t_{T}} \sim \mathcal{N}(0, I)$}. However, this formulation has no bidirectional attention between the current denoising frame $x^i$ and its history $x^{<i}$, where the strict causality forces every frame to inherit and compound the errors from its predecessors over time.

The proposed Rolling Forcing relaxes this constraint by extending the single-frame denoising window into a rolling window spanning multiple frames, as illustrated in \cref{fig:method}. Each denoising window contains consecutive frames with progressively higher noise levels in temporal order, akin to Rolling Diffusion~\citep{ruhe2024rolling}. The length of the denoising window $L_{\mathrm{win}}$ is set to the number of denoising time steps, \ie, $L_{\mathrm{win}} = T$. To ensure continuity, the next noise level of the $i$th frame is aligned with the current noise level of the $(i-1)$th frame, allowing the window to roll forward infinitely. At each roll, a clean frame is generated, and pure Gaussian noise is appended as the next frame to be synthesized. Formally, for the denoising window starting at the $i$th frame, the denoising distribution of Rolling Forcing can be defined by:
\begin{equation}
\label{eq:roll_window}
    p_\theta
    \Big(x^{i:i+T-1}_{t_{0:T-1}} \mid x^{i:i+T-1}_{t_{1:T}}, x^{<i}_0\Big) 
    = 
    \Psi\Big( G_\theta ( x^{i:i+T-1}_{t_{1:T}}, t_{1:T},  x^{<i}_0 ), t_{0:T-1}\Big),
\end{equation}
where $x^{i:i+T-1}_{t_{1:T}}$ denotes the noisy frames in the denoising window, and $x^{i:i+T-1}_{t_{0:T-1}}$ denotes the window output with each frame denoised to a lower noise level. The generator $G_\theta$ predicts clean frames conditioned on the input noisy frames, their noise levels $t_{1:T}$, and the clean history frames $x_0^{<i}$. $\Psi$ injects Gaussian noise $\epsilon_{t_{0:T-1}}$ at noise levels $t_{0:T-1}$ into the predicted clean frames, producing frames with reduced noise levels.

Since the length of the denoising window equals the number of denoising steps $T$, which is typically large (\ie, $\sim50$) in video diffusion models~\citep{wan2025wan}, the denoising window itself becomes prohibitively large. 
To manage such large windows, previous work either processes every frame independently on multiple GPUs~\citep{kim2024fifo}, or reduces $T$ to $\sim30$ using few-step samplers~\citep{xie2025progressive}.
In contrast, we adopt diffusion distillation~\citep{yin2024one, yin2024improved}, which reduces the number of denoising steps $T$ to just 5 while preserving generation quality, thereby making the denoising windows compact enough to fit on a single GPU while maintaining real-time latency.

\subsection{Temporal and Global History Context}
\label{sec:kvcache}

\begin{wrapfigure}{r}{0.48\textwidth}
\vspace{-2em}
\begin{minipage}[t]{0.48\textwidth}
  \begin{algorithm}[H]
    \caption{Rolling Forcing Training}
    \small
    \begin{algorithmic}[1]
      \Require Denoise timesteps $\{t_0, t_1, \dots, t_T\}$
      \Require Number of video frames $N$
      \Require AR diffusion model $G_\theta$ (returns KV embeddings via $G_\theta^{\mathrm{KV}}$)
      \Loop
        \State Initialize model output $\mathbf{X}_{\theta} \gets []$
        \State Initialize KV cache $\mathbf{KV} \gets []$
        \State Initialize $x^{1:T-1}_{t_{1:T-1}}$ with $G_\theta$
        \State Sample $j \sim \text{Uniform}\{0, 1, \ldots, T-1\}$
        \For{$i = 1, \dots, N$}
          \State Sample $x_{t_T}^{i+T-1} \sim \mathcal{N}(0, I)$
          \State Set $x^{i:i+T-1}_{t_{1:T}} \gets x^{i:i+T-2}_{t_{1:T-1}} \,\|\, x_{t_T}^{i+T-1}  $
          \State Select and apply RoPE to $\mathbf{KV}$ (\cref{sec:kvcache})
          \If{$i \equiv j \pmod{T}$}
            \State Enable gradient computation
            \State $\hat{x}^{i:i+T-1}_0 \gets G_\theta(x^{i:i+T-1}_{t_{1:T}}, t_{1:T},  \mathbf{KV})$
            \State $\mathbf{X}_{\theta}{\texttt{.append}}( \hat{x}^{i:i+T-1}_0)$
            \State Disable gradient computation
          \Else 
            \State $\hat{x}^{i:i+T-1}_0 \gets G_\theta(x^{i:i+T-1}_{t_{1:T}}, t_{1:T},  \mathbf{KV})$
          \EndIf
          \State $\mathbf{KV}{\texttt{.append}}( G_\theta^\text{KV}(\hat{x}^i_{0}, t_0, \mathbf{KV}))$
          \State $x^{i+1:i+T-1}_{t_{1:T-1}} \gets \Psi(\hat{x}^{i+1:i+T-1}_0, t_{1:T-1}) $
        \EndFor
        \State Update $\theta$ via DMD loss (\cref{eq:dmd})
      \EndLoop
    \end{algorithmic}
    \label{alg:training}
  \end{algorithm}
\end{minipage}
\vspace{-2em}
\end{wrapfigure}


As the clean history frames $x^{<i}_0$ accumulate during generation, handling them directly becomes computationally expensive. To address this, following~\citet{huang2025self}, we cache the key and value states of the history frames, thereby avoiding redundant recomputation when generating new frames, as illustrated in \cref{fig:method}. Note that although the attention within the denoising window is bidirectional, the attention between the frames in the denoising window and the KV cache of history frames remains causal. 
While KV caching reduces computation, the computational complexity still grows quadratically with the cache size as frames accumulate, and the cache may become large enough to cause out-of-memory errors. Given a denoising window starting at the $i$th frame $x^{i:i+T-1}_{t_{1:T}}$, we address this issue by retaining only the KV cache of the most recent $L_{\mathrm{tem}}$ history frames $x^{i-L_{\mathrm{tem}}:i-1}_0$ as temporal context to preserve short-term temporal consistency. However, relying solely on short-term history causes a gradual drift of long-range properties of the generated video (like exposure, color tone, white balance, etc.) as generation proceeds.

To maintain long-term global consistency, we cache the KV states of the initial $L_{\mathrm{glo}}$ generated frames $x_0^{1:L_{\mathrm{glo}}}$ as global context, analogous to attention sink tokens in streaming language models~\citep{xiao2023efficient}. The cache sizes $L_{\mathrm{tem}}$ and $L_{\mathrm{glo}}$ are chosen such that the total attention window size matches that of the bidirectional teacher model, \ie, $L_{\mathrm{tem}} + L_{\mathrm{glo}} + L_{\mathrm{win}} = L_{\mathrm{bidirectional}}$.
However, directly caching the initial frames leads to spilling problems. Modern video diffusion DiTs~\citep{peebles2023scalable} typically use RoPE~\citep{su2024roformer} for relative positional encoding. As the indices of the denoising frames $i:i+T-1$ increase, their relative distance to the initial cached frames grows, eventually exceeding the trained range of RoPE and producing unnatural artifacts.
To resolve this, we cache the key states of the global context frames $x_0^{1:L_{\mathrm{glo}}}$ before applying the RoPE transformation. During generation, we dynamically apply RoPE to these cached key states at the effective indices $  i - L_{\mathrm{tem}} - L_{\mathrm{glo}} :i - L_{\mathrm{tem}}-1$, treating them as being positioned immediately before the temporal context frames $x^{i-L_{\mathrm{tem}}:i-1}_0$. This adjustment preserves a fixed relative position \wrt the denoising frames, preventing excessive offsets.

\subsection{Rolling Forcing Post-Training}
\label{sec:training}

Rolling Forcing distills a pretrained bidirectional video diffusion model~\citep{wan2025wan} to a few-step causal autoregressive generator using the DMD loss (\cref{eq:dmd}). As DMD matches the holistic distribution of the entire video sequence to the data distribution $D(p_{\text{data}}(x^{1:N}) \| p_\theta(x^{1:N}))$, the calculation of the DMD loss requires a predicted clean video $\hat{x}_0^{1:N}$ during training. In SF, the predicted clean video is generated by: 
\begin{equation}
\label{eq:sf}
 \hat{x}_0^{1:N} = \Big\{ 
 \hat{x}^i_0= G_\theta(x^{i}_{t_j}, t_j, x^{<i}_0) 
 \mid i=1,2,\dots,N
 \Big\},
\end{equation}
where $j \sim \text{Uniform}\{0,1,\dots,T-1\}$ indicates each frame's noise level $t_j$ before denoising.
For Rolling Forcing, as the denoising window consists of multiple frames at different noise levels, we select the $j$th frame in each window and combine the selected frames as the predicted clean video:
\begin{equation}
\label{eq:rf_init}
    \hat{x}_0^{1:N} = \Big\{ 
    \hat{x}^i_0 = (\hat{x}^{i:i+T-1}_0)^j = 
    G_\theta(x^{i:i+T-1}_{t_{1:T}}, t_{1:T},  x^{<i}_0)^j 
    \mid i=1,2,\dots,N
    \Big\}, 
\end{equation}
where $j \sim \text{Uniform}\{0,1,\dots,T-1\}$ represents both the frame's index within the denoising window and the frame's noise level $t_j$. However, \cref{eq:rf_init} incurs $T$ times higher computational complexity than \cref{eq:sf}, because the query size is $T$ times larger. Given that DMD loss is already computationally expensive, this additional cost can easily lead to out-of-memory error even on GPUs with 80G of memory.

To address this issue, instead of backpropagating through every window (which requires gradients for each forward pass), we sample a subset of non-overlapping windows to construct the predicted clean video, as illustrated in \cref{fig:method}. Gradient computation is performed only on these selected windows, significantly reducing memory usage while retaining effective supervision. Formally, the predicted clean video is given by:
\begin{equation}
\label{eq:rf_final}
    \hat{x}_0^{1:N} = \Big\{ \hat{x}^{i:i+T-1}_0 = 
    G_\theta(x^{i:i+T-1}_{t_{1:T}}, t_{1:T},  x^{<i}_0) \mid 
    i \equiv j \pmod{T},\ 1 \leq i \leq N
    \Big\}, 
\end{equation}
where $j \sim \text{Uniform}\{0,1,\dots,T-1\}$. In each iteration, we reduce the number of forward passes requiring gradient computation from $N$ in \cref{eq:rf_init} to $\left\lceil N/T \right\rceil$\footnote{We omit the denoising windows at the start of the video for clarity in \cref{eq:rf_final,eq:roll_window}, where the window has fewer than $T$ frames. Gradient computation is still required if the window index $i$ satisfies $i \equiv j \pmod{T}$.}. The Rolling Forcing training is illustrated in \cref{alg:training}. Similar to SF, the input noisy frames $x^{i:i+T-1}_{t_{1:T}}$ during training are generated by the model rather than taken from ground truth, thus mitigating the exposure bias. However, unlike \cref{eq:sf} or \cref{eq:rf_init}, where every frame in the predicted clean video $\hat{x}_0^{1:N}$ is denoised from the same noise level $t_j$, the frames in \cref{eq:rf_final} are denoised from varying noise levels $t_{1:T}$. Consequently, frames denoised from different noise levels have different quality and clarity, leading to unnatural video $\hat{x}_0^{1:N}$ and camera movement in DMD training. To address this issue, we adopt a mixed training strategy that alternates between SF training (\cref{eq:sf}) and Rolling Forcing training (\cref{eq:rf_final}) with equal probability. The SF objective serves as a regularizer, encouraging the model to produce videos with natural camera movement. The inference adopts the Rolling Forcing paradigm alone as elaborated in \cref{alg:inference}.

\section{experiments}

\begin{table}[t]
  \small
  \setlength{\tabcolsep}{1.5pt} 
  \vspace{-2em}
  \caption{
    Comparisons with relevant baselines. We compare Rolling Forcing with representative open-source autoregressive video generation models of similar parameter sizes.
  }
  \vspace{0.5em}
  \label{tab:comparison}
  \centering
  \resizebox{\textwidth}{!}{
  \begin{tabular}{lccc|cccccc|c}
      \toprule
      \multirow{3}{*}{Model} & \multirow{3}{*}{\#Params}  & \multirow{2}{*}{Throughput} & \multirow{2}{*}{Latency} &
      \multicolumn{6}{c|}{Evaluation Scores $\uparrow$} & \multirow{3}{*}{ {\large$\Delta^{\text{Quality}}_{\text{Drift}}$} $\downarrow$ }  \\
    \cmidrule(lr){5-10}
      && (FPS) $\uparrow$ & (s) $\downarrow$ & \scriptsize Temporal   & \scriptsize Subject       & \scriptsize Background    & \scriptsize Motion    & \scriptsize Aesthetic & \scriptsize Imaging & \\
      &&                  &                  & \scriptsize Flickering  & \scriptsize Consistency   & \scriptsize Consistency   & \scriptsize Smoothness& \scriptsize Quality   & \scriptsize Quality & \\
    \midrule
    \rowcolor{catgray}
    \multicolumn{11}{l}{\textit{Diffusion Forcing Causal}}\\
    SkyReels-V2~\citep{chen2025skyreels} & 1.3B & 0.49$^{\dagger}$ & 112$^{\dagger}$ & 97.43 & 89.23 & 93.45 & 98.76 & 61.55 & 62.90 & 5.59 \\
    MAGI-1~\citep{teng2025magi}      & 4.5B & 0.19$^{\dagger}$ & 282$^{\dagger}$ & \textbf{98.21} & 90.86 & 93.25 & \textbf{99.20} & 59.91 & 59.87 & 2.15\\
    \midrule
    \rowcolor{catgray}
    \multicolumn{11}{l}{\textit{Distilled Causal}}\\
    CausVid~\citep{yin2025slow}       & 1.3B & 15.38 & 0.78 & 96.84 & 87.99 & 89.99 & 98.09 & 60.95 & 66.38 & 2.18\\
    Self Forcing~\citep{huang2025self}  & 1.3B & 15.38 & 0.78 & 97.49 & 86.48 & 90.29 & 98.47 & 60.54 & 68.68 & 1.66 \\ 
    Rolling Forcing (Ours) & 1.3B & \textbf{15.79} & \textbf{0.76} & 97.61 & \textbf{92.80} & \textbf{93.71} & 98.70 & \textbf{62.39} &	\textbf{70.75} & \textbf{0.01}\\
    \bottomrule
  \end{tabular}
  }
  \vspace{2pt}
  {\scriptsize\raggedright$^{\dagger}$ Numbers adopted from \citet{huang2025self}.\par}
  \vspace{-1em}

\end{table}

\subsection{Implementation Details}

\textbf{Model.}
We implement Rolling Forcing with Wan2.1-T2V-1.3B~\citep{wan2025wan} as our base model, which generates 5s videos at 16 FPS with a resolution of $832\times480$.
Following CausVid~\citep{yin2025slow} and Self Forcing~\citep{huang2025self}, we first initialize the base model with causal attention masking on 16k ODE solution pairs sampled from the base model.
For both ODE initialization and Rolling Forcing training, we sample text prompts from a filtered and LLM-extended version of VidProM~\citep{wang2024vidprom}. 
We set $T=5$ and perform chunk-wise denoising with each chunk containing 3 latent frames. 
The model is trained for 3,000 steps with a batch size of 8 and a trained temporal window of 27 latent frames.  
We use the AdamW optimizer for both the generator $G_\theta$ (learning rate $1.5\times10^{-6}$) and the fake score $s_{\text{gen}}$ (learning rate $4.0\times10^{-7}$). The generator is updated every 5 steps of fake score updates.

\textbf{Evaluation.} We adopt the VBench \citep{huang2024vbench} quality matrices to evaluate the generation quality over 200 randomly sampled MovieGen \citep{polyak2024movie} prompts, where the matrices measure multiple dimensions, including temporal flickering, subject consistency, background consistency, motion smoothness, aesthetic quality, and imaging quality. For fairness, all videos for quantitative evaluation are generated with the same length (30s), frame rate (16 fps), and resolution ($832 \times 480$). 
To assess quality drift in long video generation, following \citet{zhang2025packing, yin2025slow}, we compute the absolute difference in imaging quality, $\Delta^{\text{Quality}}_{\text{Drift}}$, between the first and the last 5 seconds of each video. The magnitude of $\Delta^{\text{Quality}}_{\text{Drift}}$ directly reflects the severity of error accumulation. Following \citet{huang2025self}, we evaluate real-time performance in terms of both throughput and latency. Unlike prior work that reports the first-frame latency, we measure latency after the generation process reaches a stable speed.

\begin{figure}[t]
\centering
\vspace{-1em}
\includegraphics[width=\textwidth]{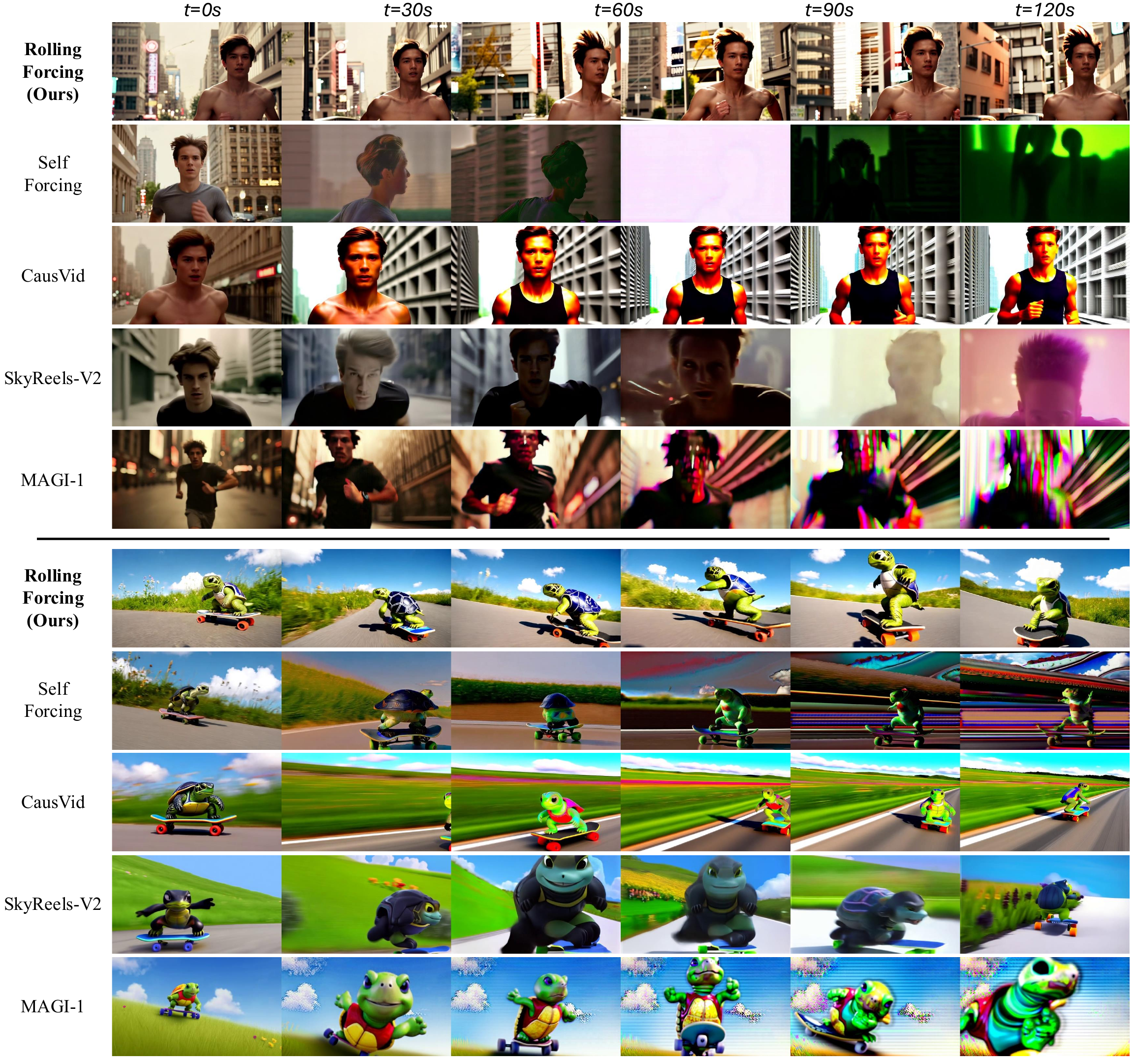}
\vspace{-1em}
\caption{Qualitative comparisons. We compare Rolling Forcing with representative open-source autoregressive video generation models on long video generation.}
\vspace{-1em}
\label{fig:comparison}
\end{figure}

\subsection{Comparisons}
We compare Rolling Forcing against several relevant open-source video generation models of comparable scale. Specifically, SkyReels-V2~\citep{chen2025skyreels} is trained under the Diffusion Forcing paradigm~\citep{chen2024diffusion}, which corrupts historical frames during inference to alleviate error accumulation. MAGI-1~\citep{teng2025magi} adopts a FIFO-style denoising paradigm~\citep{kim2024fifo} in both training and inference. We also compare against prior distillation-based approaches, including CausVid~\citep{yin2025slow} and Self Forcing~\citep{huang2025self}. Note that  SkyReels-V2, CausVid, Self Forcing, and our Rolling Forcing are all initialized from the same base model, Wan2.1-T2V-1.3B~\citep{wan2025wan}.

As shown in \cref{tab:comparison}, Rolling Forcing achieves the highest overall quality scores. In particular, it obtains a substantially lower $\Delta^{\text{Quality}}_{\text{Drift}}$, demonstrating its effectiveness in suppressing error accumulation. Qualitative comparisons in \cref{fig:comparison} further highlight that Rolling Forcing preserves high-fidelity and consistent video quality over 2 minutes of autoregressive generation, while the compared models exhibit pronounced degradation, such as color shifts, artifacts, unnatural motion, etc. In addition, Rolling Forcing achieves real-time generation with sub-second latency, marginally faster than Self Forcing and CausVid, thereby establishing its suitability for long-horizon video streaming applications.

\vspace{-0.5em}
\subsection{Ablation Studies}

\begin{figure}[t]
\centering
\vspace{-1em}
\includegraphics[width=\textwidth]{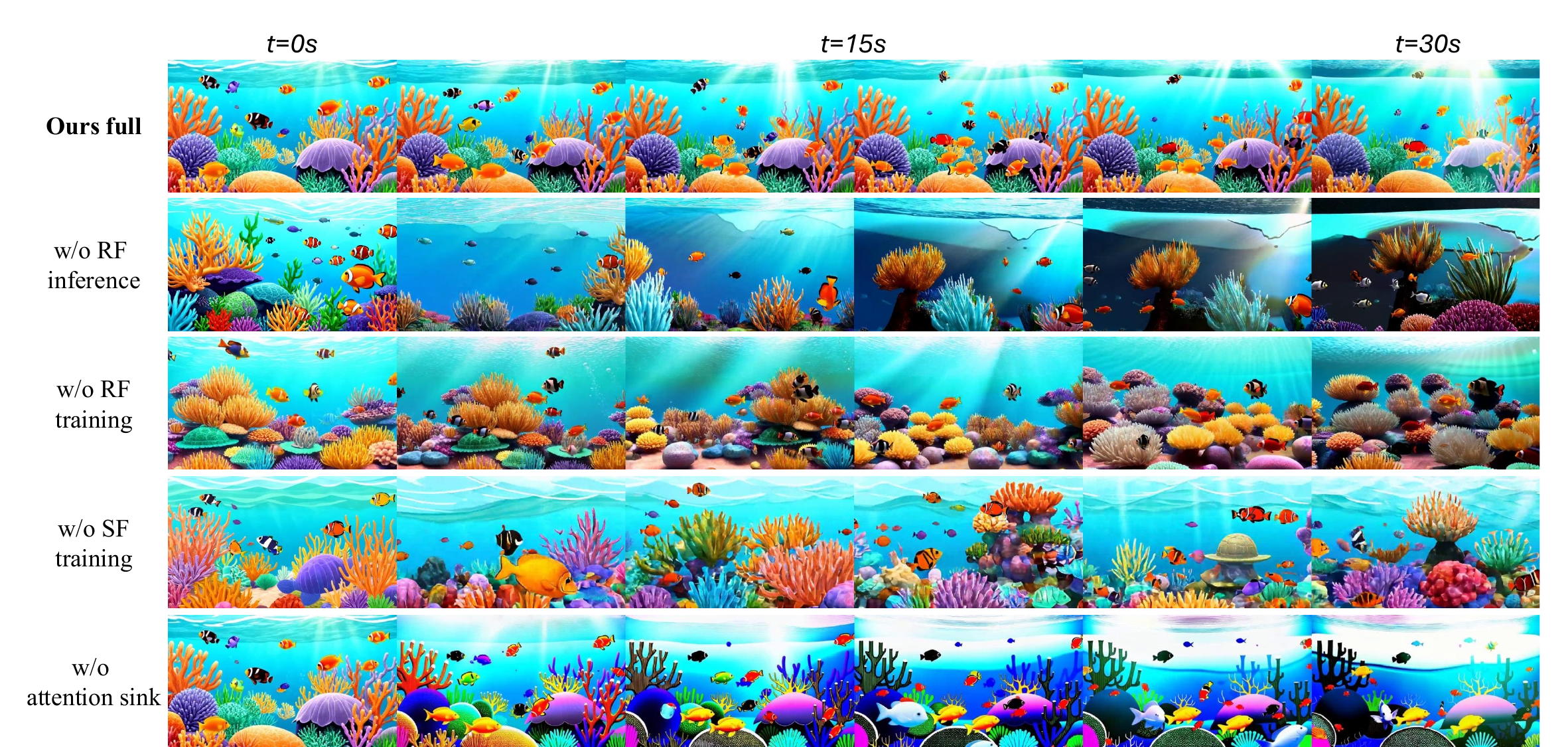}
\vspace{-1em}
\caption{Ablation studies on rolling diffusion window, mixed training strategy, and attention sink.}
\vspace{-1em}
\label{fig:ablation}
\end{figure}

\begin{wraptable}{r}{0.6\textwidth}
  \small
  \setlength{\tabcolsep}{4pt} 
  \vspace{-3em}
  \centering
  \caption{
    Ablation studies. RF refers to Rolling Forcing, and SF refers to Self Forcing.
  }
  \vspace{0.5em}
  \label{tab:ablation}
  \centering
  \resizebox{0.6\textwidth}{!}{
  \begin{tabular}{l|cccccc|c}
      \toprule
      \multirow{2}{*}{Model}  &
      \multicolumn{6}{c|}{Evaluation Scores $\uparrow$} & \multirow{2}{*}{ {\large$\Delta^{\text{Quality}}_{\text{Drift}}$} $\downarrow$ }  \\
    \cmidrule(lr){2-7}
       & \scriptsize Temp.  & \scriptsize Subj.      & \scriptsize Back.    & \scriptsize Mot.    & \scriptsize Aes. & \scriptsize Img. & \\
      \midrule
    w/o RF inference    & 95.45 & 86.01 & 89.94 & 97.36 & 57.59 & 65.19 & 5.53 \\
    w/o RF training     & 95.91 & 87.50 & 90.86 & 98.05 & 60.41 & 69.24 & 0.89 \\
    w/o SF training     & 90.83 & 83.27 & 88.14 & 95.63 & 55.30 & 62.00 & 1.62 \\
    w/o attention sink  & 97.53 & 83.22 & 87.99 & 98.56 & 58.99 & 67.30 & 4.63 \\ 
    \midrule
    Ours full           & \textbf{97.61} & \textbf{92.80} & \textbf{93.71} & \textbf{98.70} & \textbf{62.39} &	\textbf{70.75} & \textbf{0.01}\\
    \bottomrule
  \end{tabular}
  }
    \vspace{-1.5em}
\end{wraptable}

We conduct ablation studies to assess the contribution of several design options, as summarized in \cref{tab:ablation}.

\textbf{Rolling diffusion window.} We evaluate two variants: w/o RF inference and w/o RF training. In w/o RF inference, we remove the rolling denoising window and adopt frame-by-frame denoising during inference, while keeping the same training procedure and model weights as our full method. In w/o RF training, the model is trained and inferred entirely with the frame-by-frame paradigm. As shown in \cref{fig:ablation}, both variants suffer from noticeable error accumulation within 30s, demonstrating that the rolling window is crucial for suppressing long-term drift.

\textbf{Mixed training strategy.} To assess its effect, we remove the Self Forcing training objective (w/o SF training). As reported in \cref{tab:ablation}, this leads to substantial degradation in consistency and overall quality, primarily due to unnatural camera motion.

\textbf{Attention sink.} Finally, removing the global context frame (w/o attention sink) results in noticeable drift in the generated videos, as illustrated in \cref{fig:ablation}.

\section{Conclusion}

We presented \emph{Rolling Forcing}, a framework for real-time long-horizon video generation that mitigates error accumulation while sustaining sub-second latency. By introducing a rolling-window joint denoising strategy, Rolling Forcing enables mutual refinement across consecutive frames, effectively reducing long-term drift. The integration of the attention sink mechanism further enhances global consistency by anchoring initial frames as persistent context, while our efficient training algorithm enables few-step distillation over extended denoising windows while mitigating exposure bias. Extensive experiments demonstrate that Rolling Forcing achieves state-of-the-art temporal coherence and visual fidelity over multi-minute streaming sequences, significantly outperforming prior streaming approaches in both quality and efficiency.

\newpage

\bibliography{iclr2026_conference}

\begin{thebibliography}{55}
\providecommand{\natexlab}[1]{#1}
\providecommand{\url}[1]{\texttt{#1}}
\expandafter\ifx\csname urlstyle\endcsname\relax
  \providecommand{\doi}[1]{doi: #1}\else
  \providecommand{\doi}{doi: \begingroup \urlstyle{rm}\Url}\fi

\bibitem[Bansal et~al.(2024)Bansal, Bitton, Yarom, Szpektor, Grover, and Chang]{bansal2024talc}
Hritik Bansal, Yonatan Bitton, Michal Yarom, Idan Szpektor, Aditya Grover, and Kai-Wei Chang.
\newblock Talc: Time-aligned captions for multi-scene text-to-video generation.
\newblock \emph{arXiv preprint arXiv:2405.04682}, 2024.

\bibitem[Blattmann et~al.(2023{\natexlab{a}})Blattmann, Dockhorn, Kulal, Mendelevitch, Kilian, Lorenz, Levi, English, Voleti, Letts, et~al.]{blattmann2023stable}
Andreas Blattmann, Tim Dockhorn, Sumith Kulal, Daniel Mendelevitch, Maciej Kilian, Dominik Lorenz, Yam Levi, Zion English, Vikram Voleti, Adam Letts, et~al.
\newblock Stable video diffusion: Scaling latent video diffusion models to large datasets.
\newblock \emph{arXiv preprint arXiv:2311.15127}, 2023{\natexlab{a}}.

\bibitem[Blattmann et~al.(2023{\natexlab{b}})Blattmann, Rombach, Ling, Dockhorn, Kim, Fidler, and Kreis]{blattmann2023align}
Andreas Blattmann, Robin Rombach, Huan Ling, Tim Dockhorn, Seung~Wook Kim, Sanja Fidler, and Karsten Kreis.
\newblock Align your latents: High-resolution video synthesis with latent diffusion models.
\newblock In \emph{Proceedings of the IEEE/CVF conference on computer vision and pattern recognition}, pp.\  22563--22575, 2023{\natexlab{b}}.

\bibitem[Bruce et~al.(2024)Bruce, Dennis, Edwards, Parker-Holder, Shi, Hughes, Lai, Mavalankar, Steigerwald, Apps, et~al.]{bruce2024genie}
Jake Bruce, Michael~D Dennis, Ashley Edwards, Jack Parker-Holder, Yuge Shi, Edward Hughes, Matthew Lai, Aditi Mavalankar, Richie Steigerwald, Chris Apps, et~al.
\newblock Genie: Generative interactive environments.
\newblock In \emph{Forty-first International Conference on Machine Learning}, 2024.

\bibitem[Chen et~al.(2024)Chen, Mart{\'\i}~Mons{\'o}, Du, Simchowitz, Tedrake, and Sitzmann]{chen2024diffusion}
Boyuan Chen, Diego Mart{\'\i}~Mons{\'o}, Yilun Du, Max Simchowitz, Russ Tedrake, and Vincent Sitzmann.
\newblock Diffusion forcing: Next-token prediction meets full-sequence diffusion.
\newblock \emph{Advances in Neural Information Processing Systems}, 37:\penalty0 24081--24125, 2024.

\bibitem[Chen et~al.(2025)Chen, Lin, Yang, Lin, Zhu, Fan, Zhang, Chen, Chen, Ma, et~al.]{chen2025skyreels}
Guibin Chen, Dixuan Lin, Jiangping Yang, Chunze Lin, Junchen Zhu, Mingyuan Fan, Hao Zhang, Sheng Chen, Zheng Chen, Chengcheng Ma, et~al.
\newblock Skyreels-v2: Infinite-length film generative model.
\newblock \emph{arXiv preprint arXiv:2504.13074}, 2025.

\bibitem[Gu et~al.(2025)Gu, Mao, and Shou]{gu2025long}
Yuchao Gu, Weijia Mao, and Mike~Zheng Shou.
\newblock Long-context autoregressive video modeling with next-frame prediction.
\newblock \emph{arXiv preprint arXiv:2503.19325}, 2025.

\bibitem[Guo et~al.(2025)Guo, Yang, Yang, Ma, Lin, Yang, Lin, and Jiang]{guo2025long}
Yuwei Guo, Ceyuan Yang, Ziyan Yang, Zhibei Ma, Zhijie Lin, Zhenheng Yang, Dahua Lin, and Lu~Jiang.
\newblock Long context tuning for video generation.
\newblock \emph{arXiv preprint arXiv:2503.10589}, 2025.

\bibitem[Gupta et~al.(2024)Gupta, Yu, Sohn, Gu, Hahn, Li, Essa, Jiang, and Lezama]{gupta2024photorealistic}
Agrim Gupta, Lijun Yu, Kihyuk Sohn, Xiuye Gu, Meera Hahn, Fei-Fei Li, Irfan Essa, Lu~Jiang, and Jos{\'e} Lezama.
\newblock Photorealistic video generation with diffusion models.
\newblock In \emph{European Conference on Computer Vision}, pp.\  393--411. Springer, 2024.

\bibitem[Henschel et~al.(2025)Henschel, Khachatryan, Poghosyan, Hayrapetyan, Tadevosyan, Wang, Navasardyan, and Shi]{henschel2025streamingt2v}
Roberto Henschel, Levon Khachatryan, Hayk Poghosyan, Daniil Hayrapetyan, Vahram Tadevosyan, Zhangyang Wang, Shant Navasardyan, and Humphrey Shi.
\newblock Streamingt2v: Consistent, dynamic, and extendable long video generation from text.
\newblock In \emph{Proceedings of the Computer Vision and Pattern Recognition Conference}, pp.\  2568--2577, 2025.

\bibitem[Ho et~al.(2022)Ho, Chan, Saharia, Whang, Gao, Gritsenko, Kingma, Poole, Norouzi, Fleet, et~al.]{ho2022imagen}
Jonathan Ho, William Chan, Chitwan Saharia, Jay Whang, Ruiqi Gao, Alexey Gritsenko, Diederik~P Kingma, Ben Poole, Mohammad Norouzi, David~J Fleet, et~al.
\newblock Imagen video: High definition video generation with diffusion models.
\newblock \emph{arXiv preprint arXiv:2210.02303}, 2022.

\bibitem[Hong et~al.(2022)Hong, Ding, Zheng, Liu, and Tang]{hong2022cogvideo}
Wenyi Hong, Ming Ding, Wendi Zheng, Xinghan Liu, and Jie Tang.
\newblock Cogvideo: Large-scale pretraining for text-to-video generation via transformers.
\newblock \emph{arXiv preprint arXiv:2205.15868}, 2022.

\bibitem[Hu et~al.(2024)Hu, Jiang, Chen, Han, Liao, Chang, and Liang]{hu2024storyagent}
Panwen Hu, Jin Jiang, Jianqi Chen, Mingfei Han, Shengcai Liao, Xiaojun Chang, and Xiaodan Liang.
\newblock Storyagent: Customized storytelling video generation via multi-agent collaboration.
\newblock \emph{arXiv preprint arXiv:2411.04925}, 2024.

\bibitem[Huang et~al.(2025)Huang, Li, He, Zhou, and Shechtman]{huang2025self}
Xun Huang, Zhengqi Li, Guande He, Mingyuan Zhou, and Eli Shechtman.
\newblock Self forcing: Bridging the train-test gap in autoregressive video diffusion.
\newblock \emph{arXiv preprint arXiv:2506.08009}, 2025.

\bibitem[Huang et~al.(2024)Huang, He, Yu, Zhang, Si, Jiang, Zhang, Wu, Jin, Chanpaisit, et~al.]{huang2024vbench}
Ziqi Huang, Yinan He, Jiashuo Yu, Fan Zhang, Chenyang Si, Yuming Jiang, Yuanhan Zhang, Tianxing Wu, Qingyang Jin, Nattapol Chanpaisit, et~al.
\newblock Vbench: Comprehensive benchmark suite for video generative models.
\newblock In \emph{Proceedings of the IEEE/CVF Conference on Computer Vision and Pattern Recognition}, pp.\  21807--21818, 2024.

\bibitem[Jin et~al.(2024)Jin, Sun, Li, Xu, Jiang, Zhuang, Huang, Song, Mu, and Lin]{jin2024pyramidal}
Yang Jin, Zhicheng Sun, Ningyuan Li, Kun Xu, Hao Jiang, Nan Zhuang, Quzhe Huang, Yang Song, Yadong Mu, and Zhouchen Lin.
\newblock Pyramidal flow matching for efficient video generative modeling.
\newblock \emph{arXiv preprint arXiv:2410.05954}, 2024.

\bibitem[Kim et~al.(2024)Kim, Kang, Choi, and Han]{kim2024fifo}
Jihwan Kim, Junoh Kang, Jinyoung Choi, and Bohyung Han.
\newblock Fifo-diffusion: Generating infinite videos from text without training.
\newblock \emph{Advances in Neural Information Processing Systems}, 37:\penalty0 89834--89868, 2024.

\bibitem[Kodaira et~al.(2025)Kodaira, Hou, Hou, Tomizuka, and Zhao]{kodaira2025streamdit}
Akio Kodaira, Tingbo Hou, Ji~Hou, Masayoshi Tomizuka, and Yue Zhao.
\newblock Streamdit: Real-time streaming text-to-video generation.
\newblock \emph{arXiv preprint arXiv:2507.03745}, 2025.

\bibitem[Kondratyuk et~al.(2023)Kondratyuk, Yu, Gu, Lezama, Huang, Schindler, Hornung, Birodkar, Yan, Chiu, et~al.]{kondratyuk2023videopoet}
Dan Kondratyuk, Lijun Yu, Xiuye Gu, Jos{\'e} Lezama, Jonathan Huang, Grant Schindler, Rachel Hornung, Vighnesh Birodkar, Jimmy Yan, Ming-Chang Chiu, et~al.
\newblock Videopoet: A large language model for zero-shot video generation.
\newblock \emph{arXiv preprint arXiv:2312.14125}, 2023.

\bibitem[Kong et~al.(2024)Kong, Tian, Zhang, Min, Dai, Zhou, Xiong, Li, Wu, Zhang, et~al.]{kong2024hunyuanvideo}
Weijie Kong, Qi~Tian, Zijian Zhang, Rox Min, Zuozhuo Dai, Jin Zhou, Jiangfeng Xiong, Xin Li, Bo~Wu, Jianwei Zhang, et~al.
\newblock Hunyuanvideo: A systematic framework for large video generative models.
\newblock \emph{arXiv preprint arXiv:2412.03603}, 2024.

\bibitem[Li et~al.(2024)Li, Hu, Liu, Zhou, Choi, Meng, Guo, Li, Ling, and Wei]{li2024arlon}
Zongyi Li, Shujie Hu, Shujie Liu, Long Zhou, Jeongsoo Choi, Lingwei Meng, Xun Guo, Jinyu Li, Hefei Ling, and Furu Wei.
\newblock Arlon: Boosting diffusion transformers with autoregressive models for long video generation.
\newblock \emph{arXiv preprint arXiv:2410.20502}, 2024.

\bibitem[Lin et~al.(2025{\natexlab{a}})Lin, Xia, Ren, Yang, Xiao, and Jiang]{lin2025diffusion}
Shanchuan Lin, Xin Xia, Yuxi Ren, Ceyuan Yang, Xuefeng Xiao, and Lu~Jiang.
\newblock Diffusion adversarial post-training for one-step video generation.
\newblock \emph{arXiv preprint arXiv:2501.08316}, 2025{\natexlab{a}}.

\bibitem[Lin et~al.(2025{\natexlab{b}})Lin, Yang, He, Jiang, Ren, Xia, Zhao, Xiao, and Jiang]{lin2025autoregressive}
Shanchuan Lin, Ceyuan Yang, Hao He, Jianwen Jiang, Yuxi Ren, Xin Xia, Yang Zhao, Xuefeng Xiao, and Lu~Jiang.
\newblock Autoregressive adversarial post-training for real-time interactive video generation.
\newblock \emph{arXiv preprint arXiv:2506.09350}, 2025{\natexlab{b}}.

\bibitem[Lipman et~al.(2022)Lipman, Chen, Ben-Hamu, Nickel, and Le]{lipman2022flow}
Yaron Lipman, Ricky~TQ Chen, Heli Ben-Hamu, Maximilian Nickel, and Matt Le.
\newblock Flow matching for generative modeling.
\newblock \emph{arXiv preprint arXiv:2210.02747}, 2022.

\bibitem[Liu et~al.(2024)Liu, Liu, Zhou, Xu, Xie, Han, P{\'e}rez, Liu, Kahatapitiya, Jia, et~al.]{liu2024mardini}
Haozhe Liu, Shikun Liu, Zijian Zhou, Mengmeng Xu, Yanping Xie, Xiao Han, Juan~C P{\'e}rez, Ding Liu, Kumara Kahatapitiya, Menglin Jia, et~al.
\newblock Mardini: Masked autoregressive diffusion for video generation at scale.
\newblock \emph{arXiv preprint arXiv:2410.20280}, 2024.

\bibitem[Liu et~al.(2022)Liu, Gong, and Liu]{liu2022flow}
Xingchao Liu, Chengyue Gong, and Qiang Liu.
\newblock Flow straight and fast: Learning to generate and transfer data with rectified flow.
\newblock \emph{arXiv preprint arXiv:2209.03003}, 2022.

\bibitem[Long et~al.(2024)Long, Qiu, Yao, and Mei]{long2024videostudio}
Fuchen Long, Zhaofan Qiu, Ting Yao, and Tao Mei.
\newblock Videostudio: Generating consistent-content and multi-scene videos.
\newblock In \emph{European Conference on Computer Vision}, pp.\  468--485. Springer, 2024.

\bibitem[{OpenAI}(2024)]{openai_sora}
{OpenAI}.
\newblock Sora.
\newblock \url{https://openai.com/sora}, 2024.

\bibitem[Peebles \& Xie(2023)Peebles and Xie]{peebles2023scalable}
William Peebles and Saining Xie.
\newblock Scalable diffusion models with transformers.
\newblock In \emph{Proceedings of the IEEE/CVF international conference on computer vision}, pp.\  4195--4205, 2023.

\bibitem[Polyak et~al.(2024)Polyak, Zohar, Brown, Tjandra, Sinha, Lee, Vyas, Shi, Ma, Chuang, et~al.]{polyak2024movie}
Adam Polyak, Amit Zohar, Andrew Brown, Andros Tjandra, Animesh Sinha, Ann Lee, Apoorv Vyas, Bowen Shi, Chih-Yao Ma, Ching-Yao Chuang, et~al.
\newblock Movie gen: A cast of media foundation models.
\newblock \emph{arXiv preprint arXiv:2410.13720}, 2024.

\bibitem[Ruhe et~al.(2024)Ruhe, Heek, Salimans, and Hoogeboom]{ruhe2024rolling}
David Ruhe, Jonathan Heek, Tim Salimans, and Emiel Hoogeboom.
\newblock Rolling diffusion models.
\newblock In \emph{International Conference on Machine Learning}, 2024.

\bibitem[Schmidt(2019)]{schmidt2019generalization}
Florian Schmidt.
\newblock Generalization in generation: A closer look at exposure bias.
\newblock \emph{arXiv preprint arXiv:1910.00292}, 2019.

\bibitem[Seawead et~al.(2025)Seawead, Yang, Lin, Zhao, Lin, Ma, Guo, Chen, Qi, Wang, et~al.]{seawead2025seaweed}
Team Seawead, Ceyuan Yang, Zhijie Lin, Yang Zhao, Shanchuan Lin, Zhibei Ma, Haoyuan Guo, Hao Chen, Lu~Qi, Sen Wang, et~al.
\newblock Seaweed-7b: Cost-effective training of video generation foundation model.
\newblock \emph{arXiv preprint arXiv:2504.08685}, 2025.

\bibitem[Singer et~al.(2022)Singer, Polyak, Hayes, Yin, An, Zhang, Hu, Yang, Ashual, Gafni, et~al.]{singer2022make}
Uriel Singer, Adam Polyak, Thomas Hayes, Xi~Yin, Jie An, Songyang Zhang, Qiyuan Hu, Harry Yang, Oron Ashual, Oran Gafni, et~al.
\newblock Make-a-video: Text-to-video generation without text-video data.
\newblock \emph{arXiv preprint arXiv:2209.14792}, 2022.

\bibitem[Su et~al.(2024)Su, Ahmed, Lu, Pan, Bo, and Liu]{su2024roformer}
Jianlin Su, Murtadha Ahmed, Yu~Lu, Shengfeng Pan, Wen Bo, and Yunfeng Liu.
\newblock Roformer: Enhanced transformer with rotary position embedding.
\newblock \emph{Neurocomputing}, 568:\penalty0 127063, 2024.

\bibitem[Sun et~al.(2025)Sun, Wang, Li, Liu, Sun, Feng, Lao, Zhou, He, and Liu]{sun2025ar}
Mingzhen Sun, Weining Wang, Gen Li, Jiawei Liu, Jiahui Sun, Wanquan Feng, Shanshan Lao, SiYu Zhou, Qian He, and Jing Liu.
\newblock Ar-diffusion: Asynchronous video generation with auto-regressive diffusion.
\newblock In \emph{Proceedings of the Computer Vision and Pattern Recognition Conference}, pp.\  7364--7373, 2025.

\bibitem[Teng et~al.(2025)Teng, Jia, Sun, Li, Li, Tang, Han, Zhang, Zhang, Luo, et~al.]{teng2025magi}
Hansi Teng, Hongyu Jia, Lei Sun, Lingzhi Li, Maolin Li, Mingqiu Tang, Shuai Han, Tianning Zhang, WQ~Zhang, Weifeng Luo, et~al.
\newblock Magi-1: Autoregressive video generation at scale.
\newblock \emph{arXiv preprint arXiv:2505.13211}, 2025.

\bibitem[Valevski et~al.(2024)Valevski, Leviathan, Arar, and Fruchter]{valevski2024diffusion}
Dani Valevski, Yaniv Leviathan, Moab Arar, and Shlomi Fruchter.
\newblock Diffusion models are real-time game engines.
\newblock \emph{arXiv preprint arXiv:2408.14837}, 2024.

\bibitem[Wan et~al.(2025)Wan, Wang, Ai, Wen, Mao, Xie, Chen, Yu, Zhao, Yang, et~al.]{wan2025wan}
Team Wan, Ang Wang, Baole Ai, Bin Wen, Chaojie Mao, Chen-Wei Xie, Di~Chen, Feiwu Yu, Haiming Zhao, Jianxiao Yang, et~al.
\newblock Wan: Open and advanced large-scale video generative models.
\newblock \emph{arXiv preprint arXiv:2503.20314}, 2025.

\bibitem[Wang \& Yang(2024)Wang and Yang]{wang2024vidprom}
Wenhao Wang and Yi~Yang.
\newblock Vidprom: A million-scale real prompt-gallery dataset for text-to-video diffusion models.
\newblock \emph{Advances in Neural Information Processing Systems}, 37:\penalty0 65618--65642, 2024.

\bibitem[Wang et~al.(2024)Wang, Xiong, Zhou, Lin, Zhao, Kang, Feng, and Liu]{wang2024loong}
Yuqing Wang, Tianwei Xiong, Daquan Zhou, Zhijie Lin, Yang Zhao, Bingyi Kang, Jiashi Feng, and Xihui Liu.
\newblock Loong: Generating minute-level long videos with autoregressive language models.
\newblock \emph{arXiv preprint arXiv:2410.02757}, 2024.

\bibitem[Weissenborn et~al.(2019)Weissenborn, T{\"a}ckstr{\"o}m, and Uszkoreit]{weissenborn2019scaling}
Dirk Weissenborn, Oscar T{\"a}ckstr{\"o}m, and Jakob Uszkoreit.
\newblock Scaling autoregressive video models.
\newblock \emph{arXiv preprint arXiv:1906.02634}, 2019.

\bibitem[Weng et~al.(2024)Weng, Feng, Wang, Dai, Wang, Yin, Zhao, Qiu, Bao, Yuan, et~al.]{weng2024art}
Wenming Weng, Ruoyu Feng, Yanhui Wang, Qi~Dai, Chunyu Wang, Dacheng Yin, Zhiyuan Zhao, Kai Qiu, Jianmin Bao, Yuhui Yuan, et~al.
\newblock Art-v: Auto-regressive text-to-video generation with diffusion models.
\newblock In \emph{Proceedings of the IEEE/CVF Conference on Computer Vision and Pattern Recognition}, pp.\  7395--7405, 2024.

\bibitem[Xiang et~al.(2025)Xiang, Chen, Zhang, Wang, Gao, Xiang, Shang, Liu, Huang, Gao, et~al.]{xiang2025macro}
Xunzhi Xiang, Yabo Chen, Guiyu Zhang, Zhongyu Wang, Zhe Gao, Quanming Xiang, Gonghu Shang, Junqi Liu, Haibin Huang, Yang Gao, et~al.
\newblock Macro-from-micro planning for high-quality and parallelized autoregressive long video generation.
\newblock \emph{arXiv preprint arXiv:2508.03334}, 2025.

\bibitem[Xiao et~al.(2023)Xiao, Tian, Chen, Han, and Lewis]{xiao2023efficient}
Guangxuan Xiao, Yuandong Tian, Beidi Chen, Song Han, and Mike Lewis.
\newblock Efficient streaming language models with attention sinks.
\newblock \emph{arXiv preprint arXiv:2309.17453}, 2023.

\bibitem[Xie et~al.(2025)Xie, Xu, Hong, Tan, Liu, Liu, Kaufman, and Zhou]{xie2025progressive}
Desai Xie, Zhan Xu, Yicong Hong, Hao Tan, Difan Liu, Feng Liu, Arie Kaufman, and Yang Zhou.
\newblock Progressive autoregressive video diffusion models.
\newblock In \emph{Proceedings of the Computer Vision and Pattern Recognition Conference}, pp.\  6322--6332, 2025.

\bibitem[Xie et~al.(2024)Xie, Tang, Tan, Klein, Bissyand, and Ezzini]{xie2024dreamfactory}
Zhifei Xie, Daniel Tang, Dingwei Tan, Jacques Klein, Tegawend~F Bissyand, and Saad Ezzini.
\newblock Dreamfactory: Pioneering multi-scene long video generation with a multi-agent framework.
\newblock \emph{arXiv preprint arXiv:2408.11788}, 2024.

\bibitem[Yan et~al.(2021)Yan, Zhang, Abbeel, and Srinivas]{yan2021videogpt}
Wilson Yan, Yunzhi Zhang, Pieter Abbeel, and Aravind Srinivas.
\newblock Videogpt: Video generation using vq-vae and transformers.
\newblock \emph{arXiv preprint arXiv:2104.10157}, 2021.

\bibitem[Yang et~al.(2024)Yang, Zhan, Wang, Wang, Ge, Zheng, and Jin]{yang2024synchronized}
Dingyi Yang, Chunru Zhan, Ziheng Wang, Biao Wang, Tiezheng Ge, Bo~Zheng, and Qin Jin.
\newblock Synchronized video storytelling: Generating video narrations with structured storyline.
\newblock \emph{arXiv preprint arXiv:2405.14040}, 2024.

\bibitem[Yin et~al.(2024{\natexlab{a}})Yin, Gharbi, Park, Zhang, Shechtman, Durand, and Freeman]{yin2024improved}
Tianwei Yin, Micha{\"e}l Gharbi, Taesung Park, Richard Zhang, Eli Shechtman, Fredo Durand, and Bill Freeman.
\newblock Improved distribution matching distillation for fast image synthesis.
\newblock \emph{Advances in neural information processing systems}, 37:\penalty0 47455--47487, 2024{\natexlab{a}}.

\bibitem[Yin et~al.(2024{\natexlab{b}})Yin, Gharbi, Zhang, Shechtman, Durand, Freeman, and Park]{yin2024one}
Tianwei Yin, Micha{\"e}l Gharbi, Richard Zhang, Eli Shechtman, Fredo Durand, William~T Freeman, and Taesung Park.
\newblock One-step diffusion with distribution matching distillation.
\newblock In \emph{Proceedings of the IEEE/CVF conference on computer vision and pattern recognition}, pp.\  6613--6623, 2024{\natexlab{b}}.

\bibitem[Yin et~al.(2025)Yin, Zhang, Zhang, Freeman, Durand, Shechtman, and Huang]{yin2025slow}
Tianwei Yin, Qiang Zhang, Richard Zhang, William~T Freeman, Fredo Durand, Eli Shechtman, and Xun Huang.
\newblock From slow bidirectional to fast autoregressive video diffusion models.
\newblock In \emph{Proceedings of the Computer Vision and Pattern Recognition Conference}, pp.\  22963--22974, 2025.

\bibitem[Zhang \& Agrawala(2025)Zhang and Agrawala]{zhang2025packing}
Lvmin Zhang and Maneesh Agrawala.
\newblock Packing input frame context in next-frame prediction models for video generation.
\newblock \emph{arXiv preprint arXiv:2504.12626}, 2025.

\bibitem[Zhang et~al.(2025)Zhang, Bi, Hong, Zhang, Luan, Yang, Sunkavalli, Freeman, and Tan]{zhang2025test}
Tianyuan Zhang, Sai Bi, Yicong Hong, Kai Zhang, Fujun Luan, Songlin Yang, Kalyan Sunkavalli, William~T Freeman, and Hao Tan.
\newblock Test-time training done right.
\newblock \emph{arXiv preprint arXiv:2505.23884}, 2025.

\bibitem[Zhao et~al.(2024)Zhao, Liu, Wang, Chen, Wang, Chen, Zhang, and Shen]{zhao2024moviedreamer}
Canyu Zhao, Mingyu Liu, Wen Wang, Weihua Chen, Fan Wang, Hao Chen, Bo~Zhang, and Chunhua Shen.
\newblock Moviedreamer: Hierarchical generation for coherent long visual sequence.
\newblock \emph{arXiv preprint arXiv:2407.16655}, 2024.

\end{thebibliography}
\bibliographystyle{iclr2026_conference}

\newpage

\appendix

\section{Additional Implementation Details}


\begin{wrapfigure}{r}{0.48\textwidth}
\vspace{-2em}
\begin{minipage}[t]{0.48\textwidth}
  \begin{algorithm}[H]
    \caption{Rolling Forcing Inference}
    \small
    \begin{algorithmic}[1]
      \Require Denoise timesteps $\{t_0, t_1, \dots, t_T\}$
      \Require Number of video frames $N$
      \Require AR diffusion model $G_\theta$ (returns KV embeddings via $G_\theta^{\mathrm{KV}}$)

        \State Initialize model output $\mathbf{X}_{\theta} \gets []$
        \State Initialize KV cache $\mathbf{KV} \gets []$
        \State Initialize $x^{1:T-1}_{t_{1:T-1}}$ with $G_\theta$
        \For{$i = 1, \dots, N$}
          \State Sample $x_{t_T}^{i+T-1} \sim \mathcal{N}(0, I)$
          \State Set $x^{i:i+T-1}_{t_{1:T}} \gets x^{i:i+T-2}_{t_{1:T-1}} \,\|\, x_{t_T}^{i+T-1}  $
          \State Select and apply RoPE to $\mathbf{KV}$ (\cref{sec:kvcache})

          \State $\hat{x}^{i:i+T-1}_0 \gets G_\theta(x^{i:i+T-1}_{t_{1:T}}, t_{1:T},  \mathbf{KV})$
          \State $\mathbf{X}_{\theta}{\texttt{.append}}( \hat{x}^{i}_0)$

          \State $\mathbf{KV}{\texttt{.append}}( G_\theta^\text{KV}(\hat{x}^i_{0}, t_0, \mathbf{KV}))$
          \State $x^{i+1:i+T-1}_{t_{1:T-1}} \gets \Psi(\hat{x}^{i+1:i+T-1}_0, t_{1:T-1}) $
        \EndFor

    \end{algorithmic}
    \label{alg:inference}
  \end{algorithm}
\end{minipage}
\vspace{-2em}
\end{wrapfigure}


\paragraph{KV Cache.}
We configure the KV cache and denoising window sizes as $L_\text{tem}=3$, $L_\text{glo}=3$, and $L_\text{win}=15$ latent frames. When updating the KV cache with $G_\theta^\text{KV}$, the attention window attends only to recent frames, excluding the global context. This design reflects that, apart from the initial frames serving as the global context anchor, other cached frames are retained solely for preserving short-term temporal consistency. During inference, we persist the KV states of the global context frames while discarding obsolete temporal frames, thereby maintaining constant memory usage. At the start of the video, when the denoising window still includes the first frame, no temporal or global context is used. 

\paragraph{Training.}
During training, the number of generated frames $N$ is randomly sampled between 21 (the sequence length of the bidirectional teacher model) and 27 latent frames. The DMD loss is computed on the last 21 frames. Since in Wan2.1~\citep{wan2025wan} the first VAE-encoded frame is not temporally compressed and thus exhibits different statistics, we decode frames $0{:}N{-}21$ to RGB and re-encode the $(N{-}21)$-th frame into the latent space. This re-encoded frame is then concatenated with latent frames $N{-}21{:}N{-}1$ for loss computation. In this way, the first frame is only spatially compressed, ensuring consistency with the statistical distribution of the bidirectional teacher model.

\paragraph{Noise schedule and model parameterization.}
Following the Wan2.1 and Self Forcing, we adopt the flow matching framework~\citep{lipman2022flow, liu2022flow}, with time step shifting $t^\prime(k, t) = (kt / 1000) / (1 + (k-1)(t / 1000)) \cdot 1000$ and a shift factor $k=5$. The forward process is specified as $x_t = \frac{t^\prime}{1000} x + \frac{1 - t^\prime}{1000} \epsilon, \epsilon \sim \mathcal{N}(0, I)$ with $t \in [0, 1000]$.
The data prediction model is given by:
\begin{equation}
    G_\theta(x, t, c) = c_{\text{skip}} \cdot \epsilon - c_{\text{out}} \cdot v_\theta(c_{\text{in}} \cdot x_t, c_{\text{noise}}(t^\prime), c).
\end{equation}
We keep the preconditioning coefficients the same as the base models' configuration, i.e., $c_{\text{skip}} = c_{\text{in}} = c_{\text{out}} = 1$ and $c_{\text{noise}}(t) = t$.
Our few-step diffusion process employs a uniform 5-step schedule $[t_5, t_4, t_3, t_2, t_1]=[1000, 800, 600, 400, 200]$. We adopt a 5-step schedule rather than 4, as our method achieves comparable and even slightly faster generation speed than 4-step Self Forcing.

\section{VBench Scores Across All Dimensions}

\begin{table}[h]
  \small
  \setlength{\tabcolsep}{1.5pt} 
  \vspace{-1em}
  \caption{
    Full quality evaluation on VBench. 
  }
  \vspace{0.5em}
  \label{tab:quality}
  \centering
  \resizebox{\textwidth}{!}{
  \begin{tabular}{l|ccccccc|c}
      \toprule
      \multirow{2}{*}{Model}  
      &  \scriptsize Subject  & \scriptsize Background  & \scriptsize Temporal  & \scriptsize Motion   & \scriptsize Dynamic  & \scriptsize Aesthetic & \scriptsize Imaging & \scriptsize Quality \\
       & \scriptsize Consistency   & \scriptsize Consistency &  \scriptsize Flickering   & \scriptsize Smoothness & \scriptsize  Degree & \scriptsize Quality   & \scriptsize Quality & \scriptsize Score\\

    \midrule
    CausVid~\citep{yin2025slow}         &89.50&90.00&\textbf{99.41}&98.06&63.88&61.82&65.30&80.89 \\
    Self Forcing~\citep{huang2025self}  &88.61&89.53&98.90&98.57&\textbf{68.05}&60.60&68.98&81.39 \\ 
    Rolling Forcing (Ours)              &\textbf{94.80}&\textbf{95.69}&98.93&\textbf{98.63}&60.14&\textbf{62.81}&\textbf{72.31}&\textbf{84.08} \\
    \bottomrule
  \end{tabular}
  }

\end{table}

\begin{table}[h]
  \small
  \setlength{\tabcolsep}{1.5pt} 
  \vspace{-1em}
  \caption{
    Full semantic evaluation on VBench. 
  }
  \vspace{0.5em}
  \label{tab:semantic}
  \centering
  \resizebox{\textwidth}{!}{
  \begin{tabular}{l|ccccccccc|c}
      \toprule
      \multirow{2}{*}{Model}  
      &  \scriptsize Object  & \scriptsize Multiple  & \scriptsize Human  & \scriptsize \multirow{2}{*}{Color}   & \scriptsize Spatial  & \scriptsize \multirow{2}{*}{Scene} & \scriptsize Temporal & \scriptsize Appearance & \scriptsize Overall & \scriptsize Semantic \\
      
       & \scriptsize Class   & \scriptsize Objects &  \scriptsize Action   &   & \scriptsize  Relationship &    & \scriptsize Style & \scriptsize Style & \scriptsize Consistency & \scriptsize Score\\

    \midrule
    CausVid~\citep{yin2025slow}         &78.56&58.84&81.00&81.02&59.62&\textbf{31.32}&22.51&20.04&23.16&65.85 \\
    Self Forcing~\citep{huang2025self}  &80.06&62.88&\textbf{83.00}&79.80&74.76&30.59&\textbf{23.78}&\textbf{20.41}&\textbf{24.80}&69.17 \\ 
    Rolling Forcing (Ours)              &\textbf{85.92}&\textbf{64.86}&73.00&\textbf{88.16}&\textbf{78.84}&30.52&23.52&19.45&24.56&\textbf{69.78} \\
    \bottomrule
  \end{tabular}
  }

\end{table}

We conduct a comprehensive evaluation on the full VBench benchmark~\citep{huang2024vbench}, using all 946 prompts and covering all 16 metrics reported in \cref{tab:quality,tab:semantic}. For detailed metric definitions, we refer readers to the VBench paper. All values are computed with the official standardized evaluation scripts. Our method achieves substantial improvements in overall quality, particularly in frame-wise fidelity, and also outperforms distilled baselines on semantic scores.

\paragraph{Indices of the 200 sampled MovieGen prompts.} 
{\scriptsize
0, 5, 10, 15, 24, 30, 34, 38, 44, 48, 53, 60, 67, 71, 75, 79, 84, 88, 92, 98, 103, 108, 112, 116, 122, 126, 131, 137, 142, 146, 150, 157, 165, 171, 176, 182, 188, 196, 200, 207, 211, 215, 219, 225, 229, 233, 237, 242, 246, 250, 256, 261, 267, 272, 277, 284, 288, 294, 299, 303, 308, 312, 317, 321, 327, 331, 336, 340, 344, 348, 353, 357, 362, 367, 372, 376, 380, 388, 392, 396, 400, 408, 415, 423, 428, 433, 437, 441, 446, 452, 456, 460, 464, 469, 473, 477, 482, 487, 491, 495, 502, 507, 511, 515, 521, 525, 529, 533, 540, 544, 548, 553, 558, 569, 574, 578, 585, 590, 598, 602, 609, 614, 619, 626, 632, 636, 641, 647, 651, 657, 661, 666, 671, 677, 681, 686, 690, 695, 699, 704, 708, 712, 717, 722, 726, 730, 734, 739, 743, 747, 752, 756, 761, 766, 772, 776, 781, 786, 791, 795, 799, 803, 808, 812, 816, 820, 825, 829, 834, 838, 845, 849, 855, 860, 865, 870, 875, 880, 884, 888, 892, 897, 904, 908, 915, 924, 928, 933, 937, 942, 946, 954, 959, 964, 970, 976, 980, 986, 991, 996.}

\section{Interactive Video Streaming}
\begin{figure}[t]
\centering
\includegraphics[width=\textwidth]{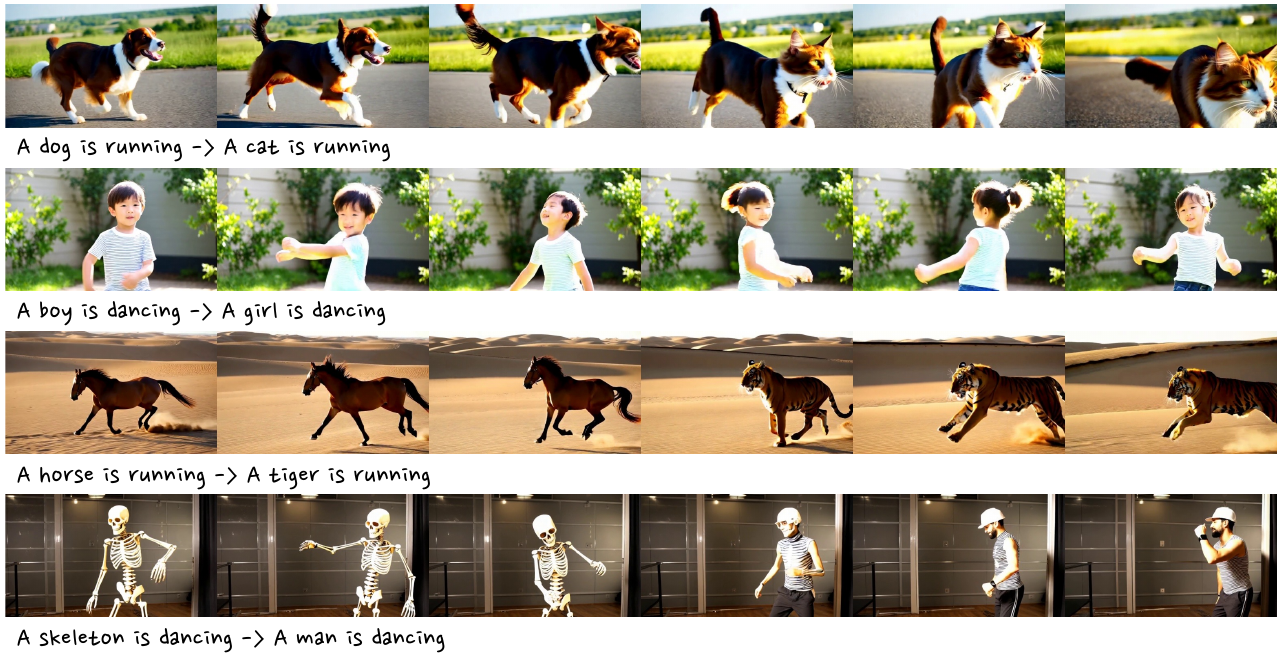}
\caption{Interactive Video Streaming. Rolling Forcing allows the users to change prompts while streaming to steer the video content.}
\label{fig:transition}
\end{figure}

In \cref{fig:transition}, we demonstrate that Rolling Forcing enables interactive video streaming, allowing users to modify prompts during generation to steer the video content. Implementing this functionality is straightforward: we discard the cross-attention cache of previous text prompts and apply the new prompts in cross-attention.

\section{Dynamic RoPE}

\begin{figure}[h]
\centering
\includegraphics[width=\textwidth]{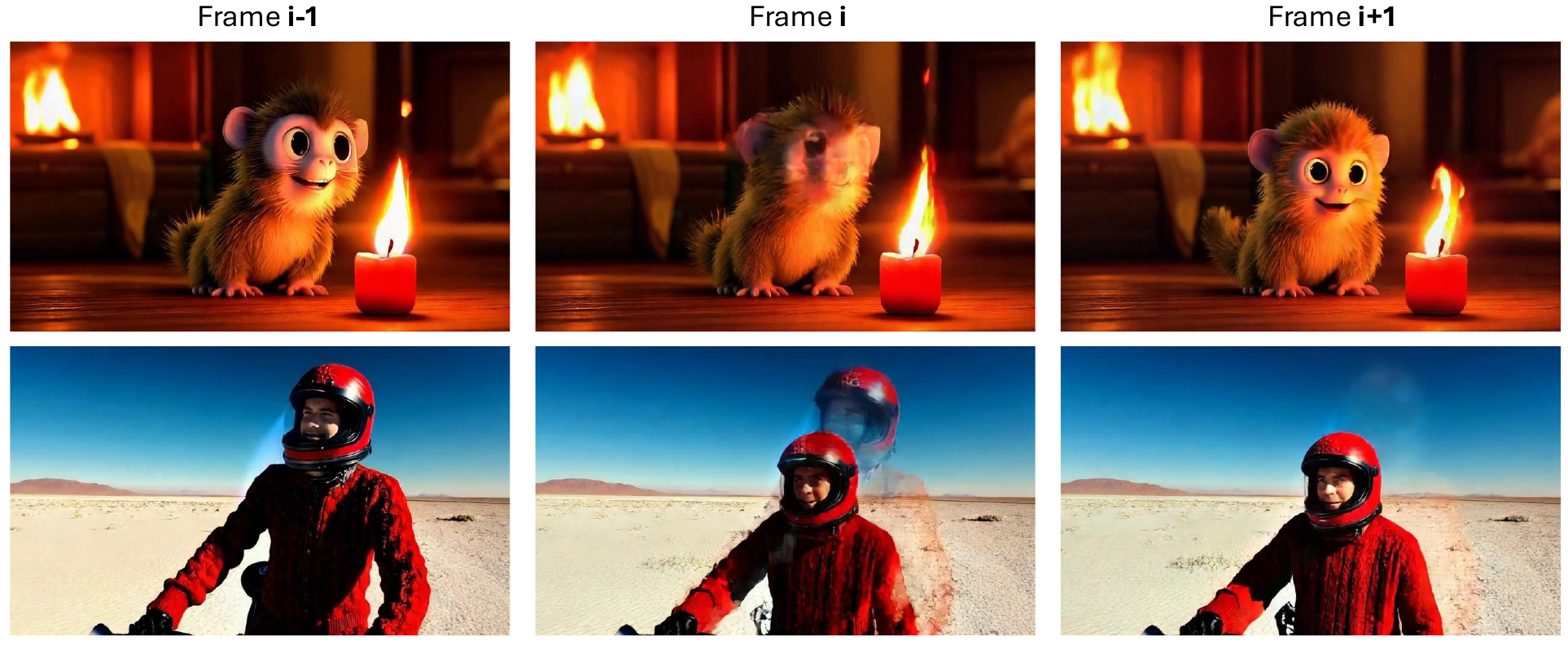}
\vspace{-1em}
\caption{Fixed RoPE indices produce strong “jumping” artifacts, where the video abruptly resets to the initial frame during streaming.}
\label{fig:jumping}
\end{figure}

The placement of RoPE indices for the global context frames is critical. Suppose the indices of the denoising window are $i{:}i+T{-}1$, and the temporal context frames are $i-L_\text{tem}{:}i{-}1$. We investigate several options for assigning indices to the global context frames:
\begin{enumerate}
\item immediately preceding the temporal context, $i-L_\text{tem}-L_\text{glo}{:}i-L_\text{tem}{-}1$ (our adopted design);

\item fixed at $0{:}L_\text{glo}{-}1$ without dynamic RoPE adjustment;

\item overlapping with the temporal context, $i-L_\text{glo}{:}i{-}1$;

\item within the denoising window, within $i{:}i+T{-}1$;

\item after the denoising window, beyond $i+T$.
\end{enumerate}
Empirically, option 2 produces strong “jumping” artifacts due to relative positions exceeding the trained offset range, as shown in \cref{fig:jumping}. Option 3 introduces flickering, as the model confuses global and temporal contexts. Option 4 collapses into static outputs, since the generated frames are forced to replicate the global context. Option 5 induces unnatural motion, as the model attempts to converge toward the misplaced global anchor. Among these, only option 1 yields consistent videos with minimal artifacts.

\section{Limitations}
While Rolling Forcing substantially suppresses error accumulation in real-time streaming video generation, several limitations remain. First, although the global context helps stabilize long-horizon consistency, frames generated in the middle are discarded once they leave the temporal context. As a result, the model retains no memory of mid-sequence content, suggesting that incorporating more advanced memory mechanisms is a promising direction for future exploration. Second, training Rolling Forcing is computationally demanding: the enlarged attention window and the DMD loss significantly increase GPU memory usage, which may limit scalability to higher-capacity models. Developing more efficient training or distillation strategies to mitigate these costs is therefore an important avenue for future work. Third, as mentioned in \citet{huang2025self}, the rolling diffusion window may increase latency in interactive applications, as future frames are partially pre-generated before the current frame is finalized. As Rolling Forcing natively supports both inference strategies, future work may consider a mixed inference strategy that interactively switches between frame-by-frame denoising during interaction and rolling denoising otherwise.

\section{Broader Societal Impact}
This work introduces real-time, long-horizon text-to-video generation, which can broaden access to interactive media, live storytelling, and educational tools by enabling continuous and responsive video synthesis. However, the ability to produce realistic long-duration content in real time also heightens risks of misuse, such as generating misleading live streams or amplifying harmful biases over extended outputs. We encourage future research to explore safeguards, including content filtering, bias mitigation, and responsible deployment practices, to ensure these capabilities are used for positive impact.

\end{document}